\documentclass{article}

 \PassOptionsToPackage{numbers,sort&compress}{natbib}
 \usepackage[final]{neurips_2025}

 \makeatletter
\providecommand{\@trackname}{}
\makeatother

 \usepackage{multirow}
 \usepackage{array}
 \usepackage{graphicx}
 \usepackage{amsmath}
\usepackage{textcomp}
\usepackage{float}
\usepackage{enumitem}

\usepackage[utf8]{inputenc} 
\usepackage[T1]{fontenc}    
\usepackage{hyperref}       
\usepackage{url}            
\usepackage{booktabs}       
\usepackage{amsfonts}       
\usepackage{nicefrac}       
\usepackage{microtype}      
\usepackage{xcolor,soul}         

\usepackage{tcolorbox}
\usepackage{booktabs}
\usepackage{makecell}
\usepackage{array}
\usepackage{graphicx}

\title{HealthSLM-Bench: Benchmarking Small Language Models for Mobile and Wearable Healthcare Monitoring}

\author{%
  Xin Wang$^{*}$,
  Ting Dang$^{*}$,
  Xinyu Zhang$^{\dagger}$,
  Vassilis Kostakos$^{*}$,
  Michael Witbrock$^{\dagger}$,
  Hong Jia$^{\dagger}$ \\
  $^{*}$School of Computing and Information Systems, University of Melbourne, Australia \\
  $^{\dagger}$School of Computer Science, University of Auckland, New Zealand \\
  \texttt{xw17@student.unimelb.edu.au}, 
  \texttt{\{ting.dang, vassilis.kostakos\}@unimelb.edu.au}, \\
  \texttt{\{xinyu.zhang, m.witbrock, hong.jia\}@auckland.ac.nz}
}

\begin{document}
\workshoptitle{The Second Workshop on GenAI for Health: Potential, Trust, and Policy Compliance}

\maketitle
\begin{abstract}
Mobile and wearable healthcare monitoring play a vital role in facilitating timely interventions, managing chronic health conditions, and ultimately improving individuals’ quality of life. Previous studies on large language models (LLMs) have highlighted their impressive generalization abilities and effectiveness in healthcare prediction tasks. However, most LLM-based healthcare solutions are cloud-based, which raises significant privacy concerns and results in increased memory usage and latency. To address these challenges, there is growing interest in compact models, Small Language Models (SLMs), which are lightweight and designed to run locally and efficiently on mobile and wearable devices. Nevertheless, how well these models perform in healthcare prediction remains largely unexplored.
We systematically benchmarked SLMs on health prediction tasks using zero-shot, few-shot, and instruction fine-tuning approaches, and deployed the best performing fine-tuned SLMs on mobile devices to evaluate their real-world efficiency and predictive performance in practical healthcare scenarios. Our results show that SLMs can achieve performance comparable to LLMs while offering substantial gains in efficiency and privacy. However, challenges remain, particularly in handling class imbalance and few-shot scenarios. These findings highlight SLMs, though imperfect in their current form, as a promising solution for next-generation, privacy-preserving healthcare monitoring.

  
\end{abstract}

\section{Introduction}

The proliferation of mobile and wearable devices, coupled with recent advances in deep learning, has significantly advanced the landscape of continuous health monitoring~\cite{dinh2019, pham2022pros, jia2024ur2m, wu2023udama}. These technologies enable a range of real-time applications, from the detection of physiological anomalies~\cite{gabrielli2025ai} to the delivery of personalized interventions~\cite{carlozzi2021careqol, he2026wearable}. Meanwhile, large language models (LLMs) have demonstrated remarkable generalization in processing heterogeneous data and performing diverse downstream tasks~\cite{ferrara2024survey, imran2024llasa}. Early studies indicate that LLM-based analysis can provide a deeper contextual interpretation of sensor data and enable more adaptive health monitoring systems compared to conventional approaches~\cite{khasentino2025personal}.

Despite this promise, major obstacles impede the practical deployment of LLM-driven wearable health solutions. Current approaches usually depend on cloud-based inference, necessitating data transmission to external servers, which raises concerns around user privacy, data security, and communication latency~\cite{das2025security, li2024privacylargelanguagemodels, xu2025camel, wang2025intelligent}. Alternatively, on-device deployment is hindered by severe resource constraints typical of mobile and wearable hardware, as well as the real-time requirements of health applications, rendering full-sized LLMs infeasible for timely inference. These challenges highlight a critical need for efficient, privacy-preserving techniques that achieve competitive performance with LLMs, while being suitable for deployment on resource-limited mobile and wearable devices. 

Small Language Models (SLMs) present a promising alternative by reducing memory consumption and facilitating deployment on mobile and wearable devices. On-device inference with SLMs not only lowers communication latency but also enhances the protection of sensitive personal data, while maintaining competitive performance on natural language processing tasks~\cite{Microsoft2023, zhang2024tinyllamaopensourcesmalllanguage, yang2024qwen2technicalreport, gemmateam2024gemma2improvingopen}. Nevertheless, their ability to interpret sensor data from mobile and wearable devices and accurately infer health conditions in real-world settings remains an open question. Although prior work~\cite{Mobicom} has demonstrated the feasibility of using SLMs on mobile devices to predict simple health status (e.g., fatigue, sleep quality), there is still a lack of comprehensive benchmarking that thoroughly evaluates SLMs for a wide range of health applications. 

To bridge this gap, we present a comprehensive benchmark, HealthSLM-Bench, which aims to evaluate a variety of state-of-the-art (SOTA) SLMs on a suit of health prediction tasks spanning three publicly available datasets. Our benchmark systematically assesses model performance using three evaluation protocols: zero-shot, few-shot, and instruction-based fine-tuning. To assess practical feasibility, we further deploy top-performing fine-tuned models on mobile devices and rigorously evaluate their on-device efficiency in terms of memory usage and inference latency. Experimental results demonstrate that SLMs can achieve comparable performance compared with SOTA healthcare LLMs across eight healthcare monitoring tasks, while substantially reducing memory and latency overheads.
Our main contributions are as follows: 
\begin{itemize} 
\item We introduce, HealthSLM-Bench, an extensive benchmark that systematically evaluates nine SOTA SLMs on eight health prediction tasks across three real-world mobile and wearable datasets. 
\item We investigate various evaluation paradigms, including zero-shot, few-shot, and instruction-based fine-tuning, providing a comprehensive performance analysis under different adaptation scenarios. 
\item We demonstrate the feasibility of deploying fine-tuned SLMs on resource-constrained mobile devices and quantify their efficiency in terms of real-world memory and latency footprints. 
\end{itemize}

\section{Related Work}
\paragraph{LLMs for health monitoring.} With the rise of mobile and wearable devices, a variety of human-centered sensing signals can be continuously collected, enabling ongoing monitoring of human health in daily life.
Recent studies have shown that the physical status data collected by mobile devices is strongly associated with health status~\cite{ballinger2018deepheart, hallgrimsson2019learning, Mullick2022}. Their work demonstrates how passive wearable sensor data can be effectively utilized to predict depression in adolescents using traditional ML models. However, these approaches, typically trained on specific datasets or tailored architectures, 
often struggle to generalize across heterogeneous tasks, and contexts~\cite{kasl2024crossstudy}. LLMs, powered by their generalization capabilities, have shown great success in the healthcare domain. For example, Health-LLM~\cite{kim2024healthllm} and MultiEEG-GPT~\cite{hu2024exploring} demonstrate the effectiveness of leveraging LLMs in healthcare monitoring through textual and physiological data. Instead of just deploying these models directly for healthcare applications, recent work has explored domain adaptation strategies such as few-shot prompting, instruction tuning, and domain-specific fine-tuning to improve performance on medical tasks~\cite{xu2024mentalllm}. Notably, Med-PaLM~\cite{singhal2023large} illustrates the benefits of combining diverse adaptation strategies (e.g. few-shot and fine-tuned) across medical datasets. Meanwhile, evaluations of GPT-4 highlight that SOTA LLMs may reduce the reliance on extensive adaptation, as they already demonstrate strong capacity for medical reasoning with limited supervision~\cite{nori2023capabilities}. More recently, applied systems such as PhysioLLM~\cite{Fang2024PhysioLLM} have integrated LLMs with wearable sensor data to provide personalized health insights, highlighting their adaptability across users and contexts. However, despite these advances, their computational overhead makes them impractical for privacy-sensitive, real-time mobile healthcare monitoring.

\paragraph{Small Language Models.}
SLMs are defined as models that are smaller in scale relative to the widely recognised LLMs, typically comprising no more than 7 billion parameters~\cite{gupta2026smalllanguagemodelsslms, hu2024minicpm}. Recent research has highlighted the efficiency and strong task performance of SLMs as lightweight alternatives to LLMs, particularly for deployment in resource-constrained environments~\cite{demystifying2025, murthy2023}. For example, Phi-3-mini-4k-Instruct, developed by Microsoft~\cite{Microsoft2023}, contains 3.8 billion parameters and is trained on a curated blend of synthetic and high-quality public datasets, emphasizing reasoning capabilities. TinyLlama-1.1B~\cite{zhang2024tinyllamaopensourcesmalllanguage} builds on Llama 2 through parameter reduction and subsequent fine-tuning using UltraChat, a broad synthetic dialogue dataset. Similarly, Google’s Gemma2-2B~\cite{gemmateam2024gemma2improvingopen}, based on Gemini research, demonstrates robust results in text generation, summarization, and reasoning benchmarks. SmolLM-1.7B from HuggingFace~\cite{huggingfacetb2024smollm} further diversifies training by leveraging synthetic educational materials and a breadth of domain samples, and Qwen2-1.5B~\cite{yang2024qwen2technicalreport} achieves SOTA performance in both coding and mathematics despite its small footprint. Meta’s Llama-3 series~\cite{grattafiori2024llama3herdmodels} continues this trend by releasing 1B and 3B parameter models designed for edge applications. While these developments affirm the viability of SLMs for a range of natural language processing tasks, the current literature leaves the open question of how effectively these compact models generalize to health prediction tasks. This is especially salient for high-stakes applications in healthcare, where accuracy and timeliness are paramount.

\paragraph{Deployment of On-Device SLMs.}  
Deploying SLMs on mobile and wearable devices is an interesting yet challenging task due to constraints such as limited computational power, memory, battery life, and the need for efficient, real-time processing. MobileAIBench~\cite{murthy2023} evaluated SLMs on an iPhone 14, offering a comprehensive framework for assessing latency, memory usage, and overall efficiency. Their results established the practical viability of running compact language models on mobile hardware. More recent research~\cite{Mobicom} explored SLMs for health event prediction~\cite{pmdata2020} in a zero-shot context, underscoring their promise as privacy-preserving and practical alternatives to LLMs for healthcare monitoring on mobile and wearable devices. Despite these advances, existing work remains limited in scope. MobileAIBench concentrated on generic NLP tasks rather than domain-specific health applications. As a result, there is a lack of systematic analysis of real-world efficiency of SLMs in health-related tasks after deployment on mobile devices. 
In comparison, our study addresses this gap by conducting comprehensive evaluations of SLMs on mobile platforms, using detailed efficiency metrics to assess their practical feasibility for mobile health monitoring applications across various datasets, model structures, and tasks.
\section{HealthSLM-Bench}
We benchmark a variety of SLMs for mobile and wearable health applications using zero-shot and few-shot learning which enables in-context learning with a limited number of task-specific examples. Additionally, we instruction-tune these models on health datasets, aiming to significantly enhance their effectiveness for healthcare monitoring tasks.

\subsection{Zero-shot and Few-shot Learning}
\begin{table*}[t!]
\centering
\caption{An Example of Prompt Construction for Zero-shot learning. $Z_S$ represents ``Zero-shot''.}
\label{table:zero-shots-prompt}
\resizebox{\columnwidth}{!}{
\begin{tabular}{>{\centering\arraybackslash}m{2cm} p{11.5cm}}
\toprule
\textbf{Context} & \textbf{Prompt} \\
\midrule

\textbf{Instruction} & You are a personalized healthcare agent trained to predict fatigue which ranges from 1 to 5 based on physiological data and user information.
\\
\midrule
\textbf{Main Query} & The recent 14-days sensor readings show: \{14\} days sensor readings show: Steps: \{``1476.0, 4809.0, ..., NaN''\} steps, Burned Calories: \{``169.0, 419.0 ..., NaN''\} calories, Resting Heart Rate: \{``53.24, 52.24, ..., 51.40''\} beats/min, Sleep Minutes: \{``110.0, 524.0, ..., 481.0''\} minutes, [Mood]: 3 out of 5. What would be the predicted fatigue level? \\
\midrule
\textbf{Output Constraints} & The predicted fatigue level is: \\
\midrule
\end{tabular}}

\begin{align}
\textit{Prompt } Z_S &= \textit{Instruction}_{Z_S} + \underline{\textit{Main query}} + \textit{Output Constraints} \tag{1}\label{eq:zs}
\end{align}
\end{table*}

\begin{table*}[t]
\centering
\caption{An Example of Prompt Construction for Few-shot learning. $Z_S$ and $F_S$ represent ``Zero-shot'' and ``Few-shot'', respectively.}
\label{table:few-shots-prompt}
\resizebox{\columnwidth}{!}{
\begin{tabular}{>{\centering\arraybackslash}m{2cm} p{11.5cm}}
\toprule
\textbf{Context} & \textbf{Prompt} \\
\midrule
\textbf{Instruction} & You are a health assistant. Your mission is to read the following examples and return your prediction based on the health query.\\[0.5em]
\midrule
\textbf{Examples} & \textless example $1$\textgreater, \textless example $2$\textgreater, ... \textless example $N$\textgreater\\
\midrule
\textbf{Question} & Finally, please answer to the below question: \textless$\textit{Prompt } Z_S$\textgreater \\ 
\bottomrule
\end{tabular}}

\begin{align}
\textit{Examples} &= (\textit{Prompt } Z_S + \textit{Answer})_{N} \tag{2} \label{eq:examples}\\
\textit{Prompt } F_S &= \textit{Instruction}_{F_S} + \textit{Examples} + \textit{Prompt } Z_S \tag{3} \label{eq:fs}
\end{align}
\vspace{-2em}
\end{table*}

\vspace{-0.1em}

\paragraph{Zero-shot learning.} 
In the zero-shot learning setting, models were evaluated without prior exposure to any example inputs during inference. Each model was provided only with a task instruction, a main query describing the 14-day summary of sensor readings, and explicit output constraints (e.g., restricting output labels for fatigue to values within the range [1–5]), as shown in Table~\ref{table:zero-shots-prompt}. This setup was designed to evaluate the intrinsic ability of the models to interpret and respond to healthcare-related queries based solely on task instructions. The zero-shot protocol thus serves as a baseline for performance, providing a reference point for subsequent experiments involving few-shot learning and instruction tuning.

\paragraph{Few-shot learning.}
Few-shot learning~\cite{brown2020languagemodelsfewshotlearners} was employed to enhance task comprehension by augmenting the model inputs with a small set of labeled examples. 
Unlike zero-shot learning, which relies solely on the model's generalized knowledge, this approach leverages in-context learning to better interpret task-specific data.
As shown in Table~\ref{table:few-shots-prompt}, the few-shot prompt (\textit{Prompt }$F_S$), formalized in Equation~\ref{eq:fs}, consists of an explicit instruction \textit{Instruction}$_{F_S}$, a set of $N$ example pairs (\textit{Prompt }$Z_S$ + \textit{Answer})$_{N}$, and the target query \textit{Prompt }$Z_S$. Specifically, the \textit{Instruction}$_{F_S}$ directs the model to review the $N$ examples before responding to the target query.
Each example follows the same structure as the zero-shot prompt, i.e., consisting of a task instruction and a main query, but also includes the corresponding answer.
This design enables the model to ground its predictions in observed input–output patterns, capturing relationships that may be less apparent in a zero-shot setting. In our experiments, we varied the number of examples $N \in \{1, 3, 5, 10\}$ to examine its impact on performance, aiming to identify the most effective configuration.
To maximize on-device efficiency, we did not implement chain-of-thought reasoning (CoT)~\cite{wei2022chain} and self-consistency (SC)~\cite{wang2023selfconsistencyimproveschainthought}, as both introduce additional token generation and computational overhead that limit practicality on resource-constrained edge devices.

\subsection{Instructional Tuning}
Instructional tuning adapts language models to follow task-specific instructions by further training them on curated instruction–response pairs~\cite{wei2022flan}. 
Unlike zero-shot or few-shot learning, which relies on a sole task description or in-context prompts at inference time, instructional tuning updates the model parameters themselves, enabling more robust and persistent task alignment.
Specifically, the instruction–response pairs were formatted using the Alpaca-style template~\cite{taori2023alpaca}, which provides a lightweight and standardized structure widely adopted in instruction-tuning benchmarks~\cite{kim2024healthllm, wang2023camels, conover2023dolly, vicuna2023open}.
To enable efficient fine-tuning, we employed Low-Rank Adaptation (LoRA)~\cite{hu2021lora}, which introduces trainable low-rank decomposition matrices into the attention and feed-forward layers while keeping the original weights frozen. LoRA is particularly well-suited for on-device inference, as it allows effective model adaptation with minimal memory and computational overhead.


\section{Experimental Setup}



\subsection{Datasets}
We evaluate our methods using three health wearable sensor datasets: PMData~\cite{pmdata2020}, GLOBEM~\cite{globem2023}, and AW-FB~\cite{dataverse2020}. From these datasets, we extract features derived from smartwatches raw sensor data, including steps, calories burned, resting heart rate, and sleep metrics, and use self-reported labels such as fatigue, stress, and readiness. For health event prediction, we format the temporal sequences of these features into 14-day windows and incorporate them into query prompts to generate predictions. The predictions produced by SLMs are then compared with the self-reported ground-truth labels. Details of each dataset are provided below. The detailed task categorization and label distribution are provided in the Appendix.

\textbf{PMData} is a dataset that integrates life-logging and activity-logging information, comprising personalized health monitoring data collected from 16 participants over a period of five months. Using the Fitbit Versa 2 smartwatch wristband~\cite{fitbitversa2}, objective signals such as calories burned, resting heart rate, step count, sleep duration, and more were gathered. In addition, participants provided self-reported measurements of their health status via the PMSys sports logging application, such as fatigue, mood, stress, etc. In our setting, these self-reports were categorized into prediction tasks with labels for fatigue, readiness, sleep quality, and stress~\cite{kim2024healthllm, Mobicom}. 

The dataset contains data relevant to the following research tasks:
\begin{itemize}
\item \textbf{Stress (STRS):} Quantification of individual stress levels, utilising both physiological measurements and self-reported data.
\item \textbf{Readiness (READ):} Evaluation of preparedness for physical exertion or exercise, based on physiological and behavioural indicators.
\item \textbf{Fatigue (FATG):} Detection and monitoring of fatigue states, as evidenced by physiological signals and self-assessment.
\item \textbf{Sleep Quality (SQ):} Comprehensive assessment of sleep quality, including metrics such as total sleep duration, sleep efficiency, and the frequency and duration of nocturnal awakenings.
\end{itemize}

\textbf{GLOBEM} is a passive sensing dataset 
for health-domain analysis. Data were gathered from 497 participants between 2018 and 2021 using a custom mobile application alongside continuous fitness tracker monitoring (24/7). This dataset captures a wide range of daily human routines, including step counts, sleep efficiency, time spent in bed after waking, time to fall asleep, and wake periods while in bed. These signals reveal associations between everyday behaviors and well-being outcomes. In our experiment, we use these behavioral signals as inputs and predict mental health conditions such as depression and anxiety~\cite{kim2024healthllm}. 

The dataset contains data relevant to the following mental health assessments:
\begin{itemize}
\item \textbf{Depression (DEP):} Detection of depressive symptoms using machine learning models that analyse user behaviour and linguistic patterns.
\item \textbf{Anxiety (ANX):} Identification of anxiety through behavioural indicators, such as disrupted sleep patterns, and physiological responses, including elevated heart rate.
\end{itemize}


\textbf{AW\_FB} is a wearable dataset designed by Harvard University to study the relationship between physical activity patterns and physiological metrics, gathered from 46 participants that wear GENEActiv~\cite{geneactiv}, Apple Watch Series 2~\cite{applewatch2} and a Fitbit Charge HR2~\cite{fitbitcharge2} in a lab-based protocol. The recorded sensor data includes daily step count, heart rate, activity duration, burned calories, and metabolic equivalent of task (MET) Value. This dataset was tested to predict 6 different physical activity intensities, including lying, sitting, walking self-paced, 3 METS, 5 METS, and 7 METS. 

The dataset contains data relevant to the following physiological and behavioural assessments:
\begin{itemize}
\item \textbf{Calorie Burn (CAL):} Estimation of individual energy expenditure during physical activities.
\item \textbf{Activity (ACT):} Classification of physical activity types based on sensor-derived data.
\end{itemize}

\subsection{Models} 
We selected nine SOTA SLMs ranging from 1B to 4B parameters, including Google’s Gemma-2-2B-it~\cite{gemmateam2024gemma2improvingopen}, Microsoft’s Phi-3-mini-4k-instruct and Phi-3.5-mini~\cite{Microsoft2023}, HuggingFace’s SmolLM-1.7B~\cite{huggingfacetb2024smollm}, Alibaba’s Qwen2-1.5B~\cite{yang2024qwen2technicalreport} and Qwen2.5-1.5B~\cite{qwen2025qwen25technicalreport}, TinyLlama’s TinyLlama-1.1B~\cite{zhang2024tinyllamaopensourcesmalllanguage}, and Meta-Llama’s Llama-3.2-1B and Llama-3.2-3B~\cite{grattafiori2024llama3herdmodels}. Further details are provided in Appendix~\ref{SLMs}.

\vspace{-0.5em}
\subsection{Implementation details}
\paragraph{Data processing.} Following previous work~\cite{kim2024healthllm, Mobicom, jia2025beyond}, we standardize all datasets into daily sequences spanning 14-day windows. Task-specific labels are assigned accordingly. Each dataset is extracted, randomly shuffled, and split into training and testing subsets in an 8:2 ratio. The tasks are categorized as either classification (fatigue, readiness, sleep quality, stress, anxiety, depression, activity) or regression (calories). The label distributions for each task are provided in the Appendix.

\paragraph{Model deployment.} To assess efficiency and feasibility, we deploy the top-performing health-domain–adapted SLMs, which is adapted for the health domain and instructional tuned using health-related datasets, on an iPhone 15 Pro Max equipped with 8 GB of RAM. 
These models are converted to the GGUF format (Generalized Graphical Unified Format)~\cite{huggingface_gguf} to ensure compatibility with lightweight inference engines such as Llama.cpp~\cite{llamacpp2023}. Due to the strict memory constraints of mobile devices, we apply 4-bit quantization to enable efficient deployment. As shown in prior studies~\cite{murthy2023}, quantization lowers computational costs while maintaining most of the model’s performance. Both the conversion and quantization steps are performed using Llama.cpp~\cite{llamacpp2023}.

\paragraph{Evaluation metrics.}
To evaluate model performance under \emph{zero-shot}, \emph{few-shot}, and \emph{instructional-tuning} settings, we use mean absolute error (MAE) for regression tasks and accuracy for classification tasks. For efficiency evaluation of mobile deployment, we assess the models latency using metrics such as Time-to-First-Token (TTFT), Input Tokens Per Second (ITPS), Output Tokens Per Second (OTPS), and Output Evaluation Time (OET) and Total Time. In addition, We also track CPU and RAM usage to evaluate on-device resource consumption. Further details are provided in the Appendix~\ref{Evaluation Metrics}.

\section{Results and Discussion}

We compare the performance of SLMs and SOTA LLMs under the same settings as in~\cite{kim2024healthllm}.

\begin{table*}[t]
\centering
\caption{Performance of LLMs and SLMs under \textbf{zero-shot (ZS)} setting across eight healthcare monitoring tasks. \textbf{STRS}: Stress, \textbf{READ}: Readiness, \textbf{FATG}: Fatigue, \textbf{SQ}: Sleep Quality, \textbf{ANX}: Anxiety, \textbf{DEP}: Depression, \textbf{ACT}: Activity, \textbf{CAL}: Calories. Best result is in \textbf{bold}, second-best result is \underline{underlined}. `-' denotes model failed to produce valid prediction.} 
\label{zero-shot performance}
\scriptsize
\setlength{\tabcolsep}{6pt}
\resizebox{\textwidth}{!}{
\begin{tabular}{llcccccccc}
\toprule
& & \multicolumn{4}{c}{\textbf{PMData}} & \multicolumn{2}{c}{\textbf{GLOBEM}} & \multicolumn{2}{c}{\textbf{AW-FB}} \\
\cmidrule(lr){3-6}\cmidrule(lr){7-8}\cmidrule(lr){9-10}
& \textbf{Model} &
\makecell{\textbf{STRS} ($\downarrow$)} &
\makecell{\textbf{READ} ($\downarrow$)} &
\makecell{\textbf{FATG} ($\uparrow$)} &
\makecell{\textbf{SQ} ($\downarrow$)} &
\makecell{\textbf{ANX} ($\downarrow$)} &
\makecell{\textbf{DEP} ($\downarrow$)} &
\makecell{\textbf{ACT} ($\uparrow$)} &
\makecell{\textbf{CAL} ($\downarrow$)} \\
\midrule

\multirow{13}{*}{\makecell{\textbf{LLMs}\\(ZS)}}
& MedAlpaca     & 0.76 & 2.18 & 46.8 & 0.68 & 1.23 & \underline{0.89} & 21.7 & 35.0 \\
& PMC\mbox{-}Llama    & 1.33 & 4.83 & 0.00 & 2.25 & 2.33 & 2.23 & -- & 43.4 \\
& Asclepius     & 0.43 & \textbf{1.44} & 27.3 & 0.45 & \textbf{0.82} & 1.10 & -- & \textbf{28.9} \\
& ClinicalCamel & 0.40 & 2.11 & 58.1 & \textbf{0.37} & 0.97 & 0.79 & 16.3 & 43.4 \\
& Flan\mbox{-}T5      & \textbf{0.36} & 1.82 & 56.8 & 0.56 & 2.84 & 2.89 & \underline{23.4} & 66.0 \\
& Palmyra\mbox{-}Med  & 0.83 & 5.01 & 43.5 & 0.44 & 2.07 & 1.99 & \textbf{29.7} & 75.3 \\
& Llama 2       & 0.57 & 2.86 & 41.2 & 0.89 & 1.19 & 1.23 & -- & -- \\
& BioMedGPT     & \underline{0.37} & 2.12 & 61.2 & \underline{0.41} & 0.95 & \textbf{0.85} & 12.2 & -- \\
& BioMistral    & 0.55 & 2.12 & 56.6 & 0.45 & \underline{0.90} & -- & 18.4 & 41.0 \\
& GPT\mbox{-}3.5       & -- & 2.38 & \underline{70.8} & 0.87 & -- & -- & 13.8 & 36.4 \\
& GPT\mbox{-}4         & -- & 2.22 & \textbf{72.2} & 0.73 & -- & -- & 22.6 & 75.2 \\
& Gemini\mbox{-}Pro    & 0.79 & \underline{1.69} & 34.0 & 0.78 & 1.03 & 0.95 & 17.7 & \underline{31.4} \\
\cmidrule{2-10}
& Mean          & 0.64 & 2.56 & 41.54 & 0.60 & 1.43 & 1.44 & 19.53 & 47.60 \\
\midrule

\multirow{10}{*}{\makecell{\textbf{SLMs}\\(ZS)}}
& Gemma-2-2b-it   & 0.72 & 2.07 & 52.84 & 0.47 & \underline{0.91} & \textbf{0.53} & -     & 105.12 \\
& Phi-3-mini-4k   & 0.46 & \textbf{1.52} & 62.88 & 0.48 & 1.08 & 1.26 & 17.39 & 93.80 \\
& SmolLM-1.7B     & 1.42 & 2.99 & 11.04 & 1.00 & 2.59 & 2.87 & \textbf{21.74} & 277.21 \\
& Qwen2-1.5B      & \textbf{0.40} & 2.03 & \underline{63.21} & \textbf{0.45} & 1.42 & 1.65 & 14.05 & 185.22 \\
& TinyLlama-1.1B  & 0.43 & 2.06 & 51.17 & 0.47 & 2.40 & 2.58 & \underline{19.73} & 198.72 \\
& Llama-3.2-1B    & \underline{0.40} & \underline{1.87} & \textbf{63.79} & 0.69 & 1.51 & 1.85 & 11.71 & 280.32 \\
& Llama-3.2-3B    & 0.67 & 2.24 & 40.80 & 0.47 & 1.26 & \underline{0.75} & 15.72 & \textbf{19.70} \\
& Phi-3.5-mini    & 0.41 & 2.34 & 61.20 & \underline{0.46} & \textbf{0.88} & 0.84 & 15.38 & \underline{56.75} \\
& Qwen2.5-1.5B    & 0.56 & 2.25 & 62.88 & 0.93 & 1.37 & 1.63 & 15.72 & 72.20 \\
\cmidrule{2-10}
& Mean & 0.61 & 2.15 & 52.20 & 0.60 & 1.49 & 1.55 & 16.40 & 143.23 \\
\bottomrule
\end{tabular}
}
\end{table*}

\subsection{Overall Performance}

\paragraph{Zero-shot learning. } As shown in Table~\ref{zero-shot performance}, SLMs achieve comparable or better performance than LLMs across the three health datasets. For stress prediction, SLMs achieve a lower mean MAE of 0.61, compared to 0.64 for LLMs, where lower values indicate better performance. SLMs also outperform LLMs in readiness and fatigue prediction, with a mean MAE of 2.15 for SLMs versus 2.56 for LLMs, and a higher mean accuracy of 52.2\% for SLMs compared to 41.54\% for LLMs. For other tasks, including sleep quality, anxiety, depression, and activity, SLMs perform within a similar range to LLMs. Among the SLMs, Gemma-2-2B-it and Phi-3-mini-4k consistently deliver strong results for fatigue and readiness, while Qwen2.5-1.5B matches or exceeds LLM performance on several tasks. However, SLMs do have some limitations. SmolLM-1.7B often underperforms relative to LLMs, and most SLMs struggle with calorie estimation, where the mean MAE is 143.23 for SLMs compared to 47.6 for LLMs, suggesting that regression tasks may be more challenging for SLMs.

\textit{In sum, under zero-shot settings, SLMs generally match or surpass LLMs on most health prediction tasks, notably achieving better results in stress, readiness, and fatigue predictions. Leading SLMs, such as Gemma-2-2B-it and Phi-3-mini-4k, show consistent strength compared with SOTA LLMs.}


\begin{table*}[t]
\centering
\caption{Performance of LLMs and SLMs under \textbf{few-shot (FS)} setting across across eight healthcare monitoring tasks. \textbf{STRS}: Stress, \textbf{READ}: Readiness, \textbf{FATG}: Fatigue, \textbf{SQ}: Sleep Quality, \textbf{ANX}: Anxiety, \textbf{DEP}: Depression, \textbf{ACT}: Activity, \textbf{CAL}: Calories. Best result is in \textbf{bold}, second-best result is \underline{underlined}. `-' denotes model failed to produce valid prediction.}
\label{few-shot performance}
\scriptsize
\setlength{\tabcolsep}{6pt}
\resizebox{\textwidth}{!}{
\begin{tabular}{llcccccccc}
\toprule
& & \multicolumn{4}{c}{\textbf{PMData}} & \multicolumn{2}{c}{\textbf{GLOBEM}} & \multicolumn{2}{c}{\textbf{AW-FB}} \\
\cmidrule(lr){3-6}\cmidrule(lr){7-8}\cmidrule(lr){9-10}
& \textbf{Model} &
\makecell{\textbf{STRS} ($\downarrow$)} &
\makecell{\textbf{READ} ($\downarrow$)} &
\makecell{\textbf{FATG} ($\uparrow$)} &
\makecell{\textbf{SQ} ($\downarrow$)} &
\makecell{\textbf{ANX} ($\downarrow$)} &
\makecell{\textbf{DEP} ($\downarrow$)} &
\makecell{\textbf{ACT} ($\uparrow$)} &
\makecell{\textbf{CAL} ($\downarrow$)} \\
\midrule

\multirow{4}{*}{\makecell{\textbf{LLMs}\\(FS-best)}}
& MedAlpaca              & \underline{0.78} & 1.94 & 36.2  & \underline{0.69} & \textbf{0.97} & \textbf{0.56} & \underline{19.3} & 36.7 \\
& GPT-3.5                & 0.94 & \textbf{1.62} & \textbf{73.9}  & 0.77 & 1.98 & 0.68 & \textbf{26.3} & \underline{26.5} \\
& GPT-4                  & \textbf{0.76} & \underline{1.64} & \underline{61.3} & \textbf{0.60} & \underline{1.11} & \underline{0.60} & 15.4 & \textbf{24.0} \\
& Gemini-Pro             & 1.10 & 2.20 & 24.8 & 0.80 & 1.30 & 1.05 & 15.0 & 37.2 \\
\cmidrule{2-10}
& Mean & 0.90 & 1.85 & 49.05 & 0.72 & 1.34 & 0.72 & 19.0 & 31.1 \\

\midrule
\multirow{9}{*}{\makecell{\textbf{SLMs}\\(FS-5)}} 
& Gemma-2-2b-it   & 0.48 & \underline{1.35} & \textbf{61.54} & \underline{0.47} & -      & -      & -      & -      \\
& Phi-3-mini-4k   & \underline{0.41} & \textbf{1.32} & \underline{57.19} & 0.49 & \underline{0.88} & \textbf{0.56} & \underline{22.10} & 37.27 \\
& SmolLM-1.7B     & -      & -      & -      & -      & \textbf{0.87} & \underline{0.76} & 17.10   & \textbf{18.58} \\
& Qwen2-1.5B      & \underline{0.41} & 1.42 & 51.51  & \textbf{0.46} & 1.20  & 1.12  & 20.40   & 29.41 \\
& TinyLlama-1.1B  & -      & -      & -      & -      & 3.15  & 3.51  & \textbf{24.10} & 37.00 \\
& Llama-3.2-1B    & 0.44  & 1.42  & 52.51  & \textbf{0.46} & 1.18  & 1.38  & 15.10   & 27.18 \\
& Llama-3.2-3B    & \underline{0.41} & 1.59 & 52.17  & \textbf{0.46} & 1.18  & 1.23  & 18.40   & 28.54 \\
& Phi-3.5-mini    & \underline{0.41} & 1.41 & 51.51  & \textbf{0.46} & 1.46  & 1.56  & \textbf{24.10} & \underline{23.69} \\
& Qwen2.5-1.5B    & \textbf{0.40} & 1.44 & 41.47  & 0.49 & 1.28  & 1.52  & 17.40   & 28.50 \\
\cmidrule{2-10}
& Mean & 0.42 & 1.42 & 52.56 & 0.47 & 1.40 & 1.45 & 19.84 & 28.77 \\
\bottomrule
\end{tabular}
}
\end{table*}

\begin{table*}[t]
\centering
\caption{Performance of LLMs and SLMs under \textbf{instruction tuning (LoRA)} setting across eight healthcare monitoring tasks. \textbf{STRS}: Stress, \textbf{READ}: Readiness, \textbf{FATG}: Fatigue, \textbf{SQ}: Sleep Quality, \textbf{ANX}: Anxiety, \textbf{DEP}: Depression, \textbf{ACT}: Activity, \textbf{CAL}: Calories. Best result is in \textbf{bold}, second-best result is \underline{underlined}. `-' denotes model failed to produce valid prediction.}
\label{lora performance}
\scriptsize
\setlength{\tabcolsep}{6pt}
\resizebox{\textwidth}{!}{
\begin{tabular}{llcccccccc}
\toprule
& & \multicolumn{4}{c}{\textbf{PMData}} & \multicolumn{2}{c}{\textbf{GLOBEM}} & \multicolumn{2}{c}{\textbf{AW-FB}} \\
\cmidrule(lr){3-6}\cmidrule(lr){7-8}\cmidrule(lr){9-10}
& \textbf{Model} &
\makecell{\textbf{STRS} ($\downarrow$)} &
\makecell{\textbf{READ} ($\downarrow$)} &
\makecell{\textbf{FATG} ($\uparrow$)} &
\makecell{\textbf{SQ} ($\downarrow$)} &
\makecell{\textbf{ANX} ($\downarrow$)} &
\makecell{\textbf{DEP} ($\downarrow$)} &
\makecell{\textbf{ACT} ($\uparrow$)} &
\makecell{\textbf{CAL} ($\downarrow$)} \\
\midrule
\multirow{2}{*}{\makecell{\textbf{LLMs}\\(lora)}} 
& HealthAlpaca-lora-7b  & 0.53 & \textbf{1.40} & 50.0 & 0.58 & \textbf{0.62} & 0.51 & 27.4 & 43.6 \\
& HealthAlpaca-lora-13b & \textbf{0.34} & 1.56 & \textbf{54.8} & \textbf{0.39} & 1.04 & \textbf{0.67} & \textbf{29.0} & \textbf{39.6} \\
\cmidrule{2-10}
& Mean & 0.44 & 1.48 & 52.4 & 0.49 & 0.83 & 0.59 & 28.2 & 41.6 \\
\midrule

\multirow{10}{*}{\makecell{\textbf{SLMs}\\(lora)}}
& Gemma-2-2b-it        & -      & -      & -    & 0.51 & 1.27 & 1.02 & \textbf{34.40} & \textbf{2.80} \\
& Phi-3-mini-4k        & \underline{0.40} & 2.14 & \underline{62.20} & 0.52 & \textbf{0.81} & 0.71 & \underline{22.40} & 9.67 \\
& SmolLM-1.7B          & 0.93  & 1.68  & 15.40 & 0.89 & 0.84 & \underline{0.54} & 16.10 & 18.87 \\
& Qwen2-1.5B           & 0.43  & 1.52  & \underline{62.20} & \underline{0.47} & 0.92 & 0.97 & 18.70 & 5.21 \\
& TinyLlama-1.1B       & \textbf{0.40} & \textbf{1.30} & \textbf{63.20} & \underline{0.47} & \underline{0.83} & 0.67 & 22.10 & 5.51 \\
& Llama-3.2-1B         & 0.43  & 2.25  & 49.80 & 0.81 & 0.86 & \textbf{0.54} & 19.20 & 5.78 \\
& Llama-3.2-3B         & 0.60  & 1.53  & 40.80 & \textbf{0.47} & 0.88 & \textbf{0.54} & 22.10 & \underline{3.64} \\
& Phi-3.5-mini         & 0.49  & 1.55  & \underline{62.20} & 0.92 & 0.88 & 0.66 & 19.40 & 12.09 \\
& Qwen2.5-1.5B         & 0.87  & \underline{1.49} & 13.00 & 0.87 & 1.04 & 0.79 & 21.70 & 4.57 \\
\cmidrule{2-10}
& Mean                 & 0.57  & 1.68  & 46.10 & 0.66 & 0.93 & 0.72 & 21.80 & 7.57 \\

\bottomrule
\end{tabular}
}
\end{table*}

\paragraph{Few-shot learning. } 
The few-shot (FS) results are shown in Table~\ref{few-shot performance}. For LLMs, we compare the best few-shot performance (FS-best) to SLMs using a range of few-shot sample counts (1, 3, 5, 10) in SLMs. As shown in Table~\ref{few-shot performance}, even when provided with in-context examples in the one-shot setting (FS-1), SLMs demonstrate competitive performance compared to their larger counterparts across multiple healthcare monitoring tasks, and also outperforms zero-shot SLMs on average.

Comparing the performance across different few-shot settings reveals interesting patterns in SLM behavior. In the FS-1 setting, SLMs achieve competitive performance levels compared to LLMs across most tasks. For instance, SLMs achieve a mean of 0.47 for stress prediction compared to LLMs' 0.90, and 0.49 for sleep quality compared to LLMs' 0.72. As the number of few-shot examples increases from FS-1 to three-shot (FS-3), five-shot (FS-5), and ten-shot (FS-10), the performance shows task-dependent variations. For stress prediction, the mean performance remains relatively stable across all few-shot settings. Similarly, sleep quality prediction maintains consistent performance throughout the different few-shot configurations.

However, certain tasks exhibit different response patterns to increased few-shot examples. Anxiety and depression prediction tasks show notable improvement as the number of examples increases from FS-1 to FS-3, with further refinement observed in subsequent settings. This suggests that mental health prediction tasks may benefit more from additional contextual examples compared to physiological monitoring tasks when using SLMs without fine-tuning, which has also been observed in recent work comparing SLMs and LLMs in mental health prediction tasks~\cite{jia2025beyond}. As shown in Table~\ref{few-shot performance}, we also observed that the collapse pattern appears at FS-1, FS-3, and FS-5, but does not occur at FS-10. This phenomenon was observed only in PMData and LifeSnaps tasks, such as stress, fatigue, sleep quality and sleep disorder, while readiness remained unaffected and no collapse was noted in GLOBEM or AW-FB tasks. Upon label distribution inspection (\textit{cf}. Appendix~\ref{Few-shot Distribution}), this trend appears to stem from limited label representation under low-shot settings, where only a few examples are provided. As the number of examples increases to FS-10, the broader label coverage yields a more representative distribution, thereby mitigating the collapse.


\textit{Overall, SLMs perform competitively with LLMs in few-shot healthcare tasks, even with just one example. More examples help models achieve more stable and reliable performance.}

\paragraph{Instruction tuning. } 


As shown in Table~\ref{lora performance}, both SLMs and SOTA LLMs~\cite{kim2024healthllm} are instruction-tuned, yet SLMs outperform LLMs in tasks such as fatigue and calorie estimation. Specifically, SLMs achieve much lower mean values for fatigue (46.1 for SLMs \textit{vs}. 52.4 for LLMs) and calorie estimation error (7.57 for SLMs compared to 41.6 for LLMs), demonstrating their superior accuracy in these important health measures. Although LLMs perform slightly better in stress, readiness, and activity prediction—with lower mean values for stress (0.44 for LLMs \textit{vs}. 0.57 for SLMs), readiness (1.48 \textit{vs}. 1.68), and higher mean values for activity (28.2 \textit{vs}. 21.8)—these differences are relatively modest compared to the clear advantages of SLMs in fatigue and calorie estimation. For other tasks such as sleep quality, anxiety, and depression, both SLMs and LLMs show similar performance, with only minor differences in mean values. Notably, SLMs like TinyLlama-1.1B and Phi-3-mini-4k stand out for their strong and consistent results across multiple tasks. In less-performing cases (e.g., activity, anxiety and sleep quality) of SLMs, we observed that SLMs tend to predict only the majority classes without attempting to predict other, weaker classes (i.e., class-imbalance bias; \textit{cf}. Appendix~\ref{Instructional tuning (LoRA) Distribution}), causing the model to become stuck at sub-optimal performance on those tasks. To mitigate this issue, we applied data augmentation (e.g., oversampling) to balance the label distribution (\textit{cf}. Appendix~\ref{app:additional_experiments}), which improved coverage of minority classes but did not lead to any performance gains.


\textit{In sum, these findings demonstrate that SLMs, when properly tuned, are not only competitive but often superior to LLMs for specific healthcare tasks, particularly fatigue and calorie estimation. This highlights the potential of SLMs for efficient, accurate, and practical healthcare applications, making them a compelling choice where resource efficiency and task-specific performance are essential.}

\subsection{Deployment Efficiency }
To investigate efficiency and computational cost in real-world deployment, we ran inference with the two top-performing models, Phi-3-mini-4k and TinyLlama-1.1B, which were instructional-tuned using LoRA, on an iPhone 15 Pro Max with 8GB memory capacity. Since the SOTA LLM HealthAlpaca-lora-7b~\cite{kim2024healthllm} did not release its checkpoint, we compared the on-device performance of selected SLMs against the baseline Llama-2-7b (the backbone of HealthAlpaca-lora-7b) using PMData to evaluate deployment efficiency. For fair comparison, we random select a total of ten samples from PMData for both Llama-2-7b, Phi-3-mini-4k and TinyLlama-1.1 to evaluate latency and hardware utilization.


As shown in Table~\ref{table:efficency}, the efficiency results of the two instruction-tuned SLMs on PMData demonstrate that SLMs preserve their latency and memory advantages over Llama-2-7b. Both TinyLlama-1.1B and Phi-3-mini-4k outperform Llama-2-7b in latency and throughput. Specifically, Phi-3-mini-4k achieves a $4.6\times$ faster time-to-first-token (TTFT) and $29\times$ faster output evaluation time (OET), with gains of over $+350\%$ in both input tokens per second (ITPS) and output tokens per second (OTPS). TinyLlama-1.1B shows even larger margins, with $21\times$ faster TTFT, $79\times$ faster OET, and more than $+2,000\%$ ITPS compared to Llama-2-7b. The memory footprint of the SLMs is also much smaller. Specifically, Phi-3-mini-4k uses $9\%$ less RAM, and TinyLlama-1.1B uses $28\%$ less than Llama-2-7b. Comparing the two SLMs, Phi-3-mini-4k offers moderate efficiency gains in some metrics but is consistently slower than TinyLlama-1.1B.


Overall, SLMs achieve substantial reductions in both input processing latency and generation latency, making them as ideal and practical solutions for resource-constrained mobile health applications.

\begin{table*}[t]
\centering
\caption{Efficiency \& Utilization of LLMs \& SLMs on the PMData dataset.}

\label{table:efficency}
\resizebox{1\textwidth}{!}{
\begin{tabular}{llccccccc}
\toprule
\textbf{} & \textbf{Model} & \textbf{TTFT(s)} & \textbf{ITPS(t/s)} & \textbf{OET(s)} & \textbf{OTPS(t/s)} & \textbf{Total Time(s)} & \textbf{CPU(\%)} & \textbf{RAM(GB)} \\
\midrule
\multirow{3}{*}{\centering\arraybackslash\begin{tabular}[c]{@{}c@{}}\textbf{PMData}\end{tabular}} 
& Phi-3-mini-4k    & 6.39  & 112.39 & 0.96  & 13.49  & 7.61  & \textbf{70.20} & 6.48 \\
& TinyLlama-1.1B   & \textbf{1.37} & \textbf{527.01} & \textbf{0.35} & \textbf{45.89} & \textbf{1.79} & 117.98 & \textbf{5.17} \\
& Llama-2-7b       & 29.12 & 24.74  & 27.85 & 3.04   & 57.43 & 379.31  & 7.15 \\
\bottomrule
\end{tabular}
}
\end{table*}

\section{Conclusion and Future Work}
In this paper, we introduce HealthSLM-Bench, a comprehensive benchmark designed to systematically evaluate SOTA SLMs on healthcare monitoring tasks under zero-shot, few-shot, and instruction-tuning scenarios. Furthermore, we assess the efficiency of these models following instruction-tuning through on-device deployment experiments. Our study shows that SLMs can match or even surpass much larger LLMs after adapted with few-shot and instructional tuning while delivering superior efficiency gain, making them practical for real-time on-device deployment. At the same time, we also identified their limitations in few-shot prompting and restricted effectiveness in instruction tuning, particularly under class-imbalanced datasets.
Both limitations point to several promising directions for future work. The first is to investigate the underlying causes of the few-shot anomaly and explore robust prompt design to prevent collapse. Another direction is to explore imbalance-aware training approaches, for example by adjusting loss weighting or augmenting minority-class samples, to reduce class bias during SLM fine-tuning. Additionally, leveraging adaptive techniques such as test-time adaptation~\cite{jia2024tinytta} could further strengthen SLM generalisation in health applications. Taken together, our benchmark establishes SLMs as a promising yet imperfect solution for efficient and privacy-preserving healthcare applications, motivating further exploration to address these challenges.





\bibliographystyle{unsrtnat}
\bibliography{reference}
\clearpage
\appendix
\section*{Appendix}

\section{Implementation Details}
We fine-tune our SLMs on a NVIDIA A100 80GB GPUs with a batch size of 128 with 3 number of epochs for the purpose of fine-tuning, with Adam optimizer and a learning rate as 5e-5 (cosine learning rate scheduler and dynamic warmup steps of 5\% of dataset size). It took about 7 hours for 9 SLMs in 3 epochs of training with the default training setting. We adopt greedy decoding method with sampling set to False. We utilize the same prompt of zero-shot for LoRA tuned SLMs inference. To ensure re-productiveness, we employ the greedy decoding strategy to make the output prediction deterministic. While most language models default to sampling-based decoding (e.g., top-$k$, top-$p$), we explicitly disabled these strategies to maintain reproducibility across runs. To better simulate edge-device conditions, where computational resources are constrained, we capped the maximum number of generated tokens at 30. Generation stops once this limit is reached, even if the answer is incomplete, which balances efficiency and response quality. The codes and fine-tuned models will be made publicly available upon the release of the camera-ready version of this paper.


\section{Additional Experiments}
\label{app:additional_experiments}
To solve the class-imbalance issue we observed under instructional-tuning (LoRA), we employed a data augmentation method (oversampling) to balance minority classes in the training data. Specifically, samples from underrepresented classes were randomly duplicated until all classes reached the size of the majority class. In the fine-tuning setup, we specifically apply label-grouping prepossessing strategy, which groups samples by their labels instead of arranging them in a random order. This grouping ensures that each mini-batch contains samples with a consistent label, stabilizing gradient updates and improving class representation during fine-tuning. For better assess the effectiveness of class-imbalance mitigation, model performance was further evaluated using the Macro F1-score, in addition to Accuracy and MAE. 
\begin{table*}[h]
\centering 
\caption{LoRA performance of SLMs fine-tuned on augmented datasets (PMData \& GLOBEM) compared to those fine-tuned on the original dataset (with label-grouping). The best results are shown in \textbf{bold}, and the second-best results are \underline{underlined}. ``OR'' denotes fine-tuning with the original dataset, while ``OS'' denotes fine-tuning on the oversampled datasets.}

\label{tab: data-augmentation}
\scriptsize
\resizebox{0.92\textwidth}{!}{
\begin{tabular}{llcccccc}
\toprule
& \textbf{Model} & \textbf{STRS ($\downarrow$)} & \textbf{READ ($\downarrow$)} & \textbf{FATG ($\uparrow$)} & \textbf{SQ ($\downarrow$)} & \textbf{ANX ($\downarrow$)} & \textbf{DEP ($\downarrow$)} \\
\midrule
\multirow{2}{*}{\centering\arraybackslash\begin{tabular}[c]{@{}c@{}}\textbf{LLMs}\end{tabular}} 
& HealthAlpaca-lora-7b  & 0.53 & \textbf{1.40} & 50.0 & 0.58 & \textbf{0.62} & 0.51 \\
& HealthAlpaca-lora-13b & \textbf{0.34} & 1.56 & \textbf{54.8} & \textbf{0.39} & 1.04 & \textbf{0.67} \\
\cmidrule{2-8}
& Mean & 0.44 & 1.48 & 52.4 & 0.49 & 0.83 & 0.59 \\

\midrule
\multirow{9}{*}{\centering\arraybackslash\begin{tabular}[c]{@{}c@{}}\textbf{SLMs} \\(OR)\end{tabular}} 
& gemma-2-2b-it         & -     & -     & -     & -     & 1.064 & 0.576 \\
& Phi-3-mini-4k         & \underline{0.395} & 1.997 & 62.5 & 0.468 & \textbf{0.806} & 0.545 \\
& SmolLM-1.7B           & 0.526 & 1.753 & 12.4 & 0.900 & 0.890 & \underline{0.542} \\
& Qwen2-1.5B            & \textbf{0.388} & \textbf{1.304} & \underline{63.2} & \underline{0.462} & 0.940 & 0.933 \\
& TinyLlama-1.1B        & 0.398 & \underline{1.311} & \textbf{63.5} & 0.475 & 0.876 & 0.555 \\
& Llama-3.2-1B          & 0.415 & 2.003 & 52.8 & \textbf{0.448} & 0.866 & \textbf{0.535} \\
& Llama-3.2-3B          & 0.501 & 1.525 & 38.1 & 0.495 & 0.882 & \textbf{0.535} \\
& Phi-3.5-mini          & 0.445 & 1.642 & 62.2 & 0.957 & \underline{0.833} & 0.766 \\
& Qwen2.5-1.5B          & -     & 1.354 & 49.1 & 0.720 & -     & - \\
\cmidrule{2-8}
& Mean & 0.438 & 1.611 & 50.5 & 0.616 & 0.895 & 0.623 \\

\midrule
\multirow{9}{*}{\centering\arraybackslash\begin{tabular}[c]{@{}c@{}}\textbf{SLMs} \\(OS)\end{tabular}} 
& gemma-2-2b-it     & \textbf{0.48} & \underline{1.73} & 41.70 & \underline{0.48} & 1.98 & 1.94 \\
& Phi-3-mini-4k     & 1.46 & 3.70 & 15.4 & 1.79 & 1.01 & 0.92 \\
& SmolLM-1.7B       & 1.03 & 1.85 & 21.4 & 1.21 & 1.68 & 1.95 \\
& Qwen2-1.5B        & 1.68 & 3.32 & 0.7 & 0.80 & 2.82 & 2.84 \\
& TinyLlama-1.1B    & 0.61 & 4.80 & \textbf{58.9} & 0.64 & 3.48 & 1.15 \\
& Llama-3.2-1B      & 1.01 & 7.02 & 31.8 & 0.90 & \textbf{0.84} & \textbf{0.61} \\
& Llama-3.2-3B      & 0.75 & 3.31 & 29.8 & \textbf{0.48} & 1.03 & 1.25 \\
& Phi-3.5-mini      & \underline{0.55} & 2.45 & \underline{50.8} & 0.73 & \underline{0.92} & \underline{0.84} \\
& Qwen2.5-1.5B      & 1.08 & \textbf{1.55} & 4.4 & 1.00 & 1.46 & 1.72 \\
\cmidrule{2-8}
& Mean   & 0.96 & 3.30 & 28.3 & 0.89 & 1.69 & 1.47 \\

\bottomrule
\end{tabular}}
\end{table*}

\begin{table*}[t]
\centering
\caption{LoRA Performance of 9 SLMs fine-tuned with augmented dataset compared to finetuned with original. All tasks are evaluated by Macro F1-score, which higher value indicates better performance. Best results in \textbf{bold}, second best are \underline{underlined}. ``-'' denotes invalid or unreasonable responses generated by models, while ``N/A'' indicates the result is not available in original paper.}
\label{tab: data-augmentation-f1}
\scriptsize
\resizebox{0.92\textwidth}{!}{
\begin{tabular}{lccccccc}
\toprule
& \textbf{Model} & \textbf{STRS ($\uparrow$)} & \textbf{READ ($\uparrow$)} & \textbf{FATG ($\uparrow$)} & \textbf{SQ ($\uparrow$)} & \textbf{ANX ($\uparrow$)} & \textbf{DEP ($\uparrow$)} \\
\midrule
\multirow{2}{*}{\centering\arraybackslash\begin{tabular}[c]{@{}c@{}}\textbf{LLMs}\end{tabular}} 
& HealthAlpaca-lora-7b   & N/A & N/A & 19.0 & N/A & N/A & N/A \\
& HealthAlpaca-lora-13b  & N/A & N/A & \textbf{45.0} & N/A & N/A & N/A \\
\midrule
\multirow{9}{*}{\centering\arraybackslash\begin{tabular}[c]{@{}c@{}}\textbf{SLMs} \\(OR)\end{tabular}} 
& gemma-2-2b-it         & -    & -   & -    & -    & 14.8 & 15.1 \\
& Phi-3-mini-4k         & 12.9 & 2.4 & 15.4 & 15.4 & 13.2 & \textbf{17.8} \\
& SmolLM-1.7B           & 12.6 & 5.0 & 6.0  & 11.6 & \textbf{19.6} & 16.9 \\
& Qwen2-1.5B            & \textbf{14.3} & 4.0 & 15.5 & 18.1 & 12.0 & 14.3 \\
& TinyLlama-1.1B        & \underline{13.7} & 3.9 & \underline{17.5} & 14.2 & \underline{16.0} & 17.2 \\
& Llama-3.2-1B          & 12.7 & \textbf{7.5} & \textbf{18.4} & \textbf{30.2} & 14.0 & 15.3 \\
& Llama-3.2-3B          & \textbf{14.3} & \underline{6.9} & 17.1 & 21.8 & 12.3 & 15.3 \\
& Phi-3.5-mini          & 12.4 & 5.1 & 15.4 & 8.9  & 15.2 & \underline{17.4} \\
& Qwen2.5-1.5B          & -    & 6.2 & 13.6 & \underline{23.2} & -   & - \\
\cmidrule{2-8}
& Mean       & 13.3 & 5.1  & 14.9 & 17.9 & 14.6 & 16.2 \\

\midrule
\multirow{9}{*}{\centering\arraybackslash\begin{tabular}[c]{@{}c@{}}\textbf{SLMs} \\(OS)\end{tabular}} 
& gemma-2-2b-it         & \textbf{24.8} & \textbf{6.3} & 17.5 & \textbf{32.6} & 7.4 & 8.7 \\
& Phi-3-mini-4k         & 8.7           & 3.3          & 6.7  & 7.0           & 15.5 & 12.0 \\
& SmolLM-1.7B           & 16.7          & 5.0          & 16.1 & 12.0          & 10.8 & 7.5 \\
& Qwen2-1.5B            & 1.4           & 2.9          & 0.3  & 19.6          & 6.4  & 4.0 \\
& TinyLlama-1.1B        & 16.3          & 1.3          & \underline{19.4} & 22.7 & 5.9  & 9.9 \\
& Llama-3.2-1B          & 6.9           & 3.6          & 12.3 & 18.1          & \textbf{21.8} & \textbf{19.5} \\
& Llama-3.2-3B          & \underline{18.8} & \underline{5.3}  & 15.0 & 19.0  & \underline{17.4} & \underline{17.6} \\
& Phi-3.5-mini          & 14.8          & 2.1          & \textbf{19.6} & \underline{29.9} & 15.6 & 11.6 \\
& Qwen2.5-1.5B          & 4.3           & 3.8          & 2.0  & 8.0           & 7.7  & 2.9 \\
\cmidrule{2-8}
& \textbf{Mean} & 12.5 & 3.7 & 12.1 & 18.8 & 12.1 & 10.4 \\

\bottomrule
\end{tabular}}
\end{table*}

\paragraph{Performance comparison (Accuracy, MAE). }{As shown in Table~\ref{tab: data-augmentation}, SLM performance generally declined when fine-tuned on oversampled datasets (OS) compared with the original datasets (OR). For instance, the mean error of STRS, READ, SQ, ANX, and DEP increased from 0.438, 1.611, 0.616, 0.805, and 0.623 to 0.961, 3.303, 28.3, 0.892, and 1.691, respectively, while the mean accuracy of FATG decreased from 48.0 to 28.3. When comparing the best results, SLMs fine-tuned on OS also showed a similar decline, though the degradation was less pronounced than in the mean performance. Specifically, the lowest MAE of STRS and READ increased from 0.388 and 1.304 to 0.482 and 1.549, respectively (a similar trend was observed for SQ, ANX, and DEP), while the highest accuracy of FATG dropped from 63.5 to 58.9. For some models, such as Qwen2-1.5B and Qwen2.5-1.5B, exhibited the most severe degradation, with FATG accuracies of only 0.7\% and 4.4\%, compared to 63.2\% and 49.1\% fine-tuned under the original datasets. These observations indicate that strategies like oversampling tends to amplify existing noise in minority classes, leading to reduced generalization under instruction-tuning. }

\paragraph{Analysis on distribution inspection. }{
Upon inspection, as shown in Figure~\ref{fig: Predicted distributions of SLMs (LoRA) on PMData, comparing models fine-tuned on the original and oversampled datasets.}, SLMs fine-tuned on oversampled datasets did demonstrate improved coverage of minority classes compared to those fine-tuned on the original datasets. Models, such as \textit{gemma-2-2b-it} and \textit{phi-3.5-mini}, began predicting a broader range of labels on STRS, FATG, and SQ, bringing their frequency distributions closer to the true distribution, where labels 2–4 are the most representative. Similarly, on the GLOBEM dataset (Figure~\ref{fig: Predicted distributions of SLMs (LoRA) on GLOBEM, comparing models fine-tuned on the original and oversampled datasets.}), the best-performing models (\textit{Llama-3.2-1B} and \textit{Phi-3.5-mini}) expanded their predictions beyond dominant labels (0, 1) to include rare labels such as 3 and 4. 
This improvement indicates that oversampling mitigates single-class dominance and enables SLMs to capture a more balanced label distribution. However, residual mismatches still persist. In particular, \textit{Qwen2.5-1.5B} tends to over-predict label 4 on PMData tasks and label 2 on GLOBEM tasks, suggesting that oversampling amplifies noise in underrepresented classes and thereby explains the degradation in performance observed earlier. }

\paragraph{Performance comparison (Macro F1-Score). }{As a more reliable measure of performance on these class-imbalanced datasets, the Macro F1-scores reported in Table~\ref{tab: data-augmentation-f1} further support our earlier observations on the predicted distributions. In particular, SLMs fine-tuned on oversampled datasets achieved higher F1-scores than those trained on the original datasets across most tasks, consistent with their predicted distributions being closer to the true label distribution. For example, on PMData tasks such as STRS, FATG, and SQ, the top-performing models obtained F1-scores of 24.8, 19.6, and 32.6 trained on oversampled data, compared to 14.3, 18.4, and 30.2 trained on original data. Similarly, on GLOBEM tasks (ANX, DEP), SLMs trained with oversampled datasets achieved the best F1-scores of 21.8 and 19.5, surpassing 19.6 and 17.8 trained with original datasets. These results confirm that oversampling enhances class diversity and balances predictions, even though overall accuracy and MAE slightly decline. However, as oversampling simply duplicates existing data, it cannot introduce new variability, which may cause models to overfit to minority patterns and limit generalization on unseen data.}

\textit{In sum, oversampling improves class balance and prediction diversity for SLMs, as reflected by higher Macro F1-scores and broader label coverage. However, these gains come at the cost of reduced performance, likely due to overfitting on duplicated samples and weaker generalization. 
These observations highlight oversampling as an effective yet imperfect strategy for mitigating class imbalance in SLM predictions, suggesting that future work should explore more advanced augmentation methods that maintain its stability while increasing data diversity.}

\section{Small Language Models}
\label{SLMs}
We selected 9 most state-of-the-art SLMs between 1 to 4B from top-tier tech companies. The details of each SLMs are listed below:
\begin{itemize}[leftmargin=*]
\item \textbf{Phi-3-mini-4k-Instruct}~\cite{microsoft2024phi3mini4kinstruct}: Microsoft's smallest model in the Phi-3 family. It has 3.8 billion parameters, trained on a combination of synthetic data and selected publicly available website data, with an emphasis on high-quality and reasoning-dense properties.
\item \textbf{Phi-3.5-mini-Instruct}~\cite{microsoft2024phi3mini4kinstruct}: A upgrade version of phi-3-mini-4k-instruct. It is built in the same architecture and dataset upon phi-3, but trained with a focus on reasoning dense data for better instruction alignment and multi-step reasoning.
\item \textbf{TinyLlama-1.1B-Chat-v1.0}~\cite{tinylama2024tinyllama}: Distilled version of Llama 2. It uses the same architecture and tokenizer as LLaMA but is compact with 1.1 billion parameters. It was fine-tuned on the UltraChat dataset (contains field-cross synthetic dialogues generated by ChatGPT), making it compatible with a wide range of tasks.
\item \textbf{Gemma2-2B-it}~\cite{google2024gemma2}: Google’s SOTA open-source model, built on the same research and technology as the Gemini models but scaled down to 2 billion parameters. It is well-suited for text generation tasks such as question answering, summarization, and reasoning.
\item \textbf{SmolLM-1.7B-Instruct}~\cite{huggingfacetb2024smollm}: HuggingFace's flagship model, it has 1.7 billion parameters and is trained on SmolLM-Corpus which consists of synthetic textbooks, stories, and educational Python and web samples.
\item \textbf{Qwen2-1.5B-Instruct}~\cite{qwen2024qwen2}: Ailibaba's state-of-the-art SLM in Qwen2 family. It has only 1.5 billion parameters and is trained on diverse instruction-followed tasks. The included coding and mathematics data for training makes it perform well in coding and quantitative reasoning tasks.

\item \textbf{Qwen2.5-1.5B-Instruct}~\cite{qwen2025qwen25technicalreport}: An upgraded version of Qwen2. It is built on the same dataset and architecture, but places greater emphasis on coding and mathematics tasks, making it more optimized for reasoning and math.

\item \textbf{Llama-3.2-1B-Instruct}~\cite{meta2024llama32modelcard}: Meta-llama's state-of-the-art SLM. It shares the identical architecture and pre-trained datasets upon Llama3, but is compressed to 1B parameters.
\item \textbf{Llama-3.2-3B-Instruct}~\cite{meta2024llama32modelcard}: 3B version of Llama-3.2-1B-Instruct.
\end{itemize}

\section{Task categorization and Label Distribution}
\label{datasets}

\subsection{PMData}
 \begin{itemize}[leftmargin=*]
    \item Stress (STRS): Estimation of an individual’s stress level based on physiological data and self-reported measures. (0-5, Classification)
    \item Readiness (READ): Assessment of an individual’s readiness for physical activity/exercise. (0-10, Classification)
    \item Fatigue (FATG): Monitoring of signs of tiredness or exhaustion based on sports and life-log data in the last 14 days. (1-5, Classification)
    \item Sleep Quality (SQ): Estimation of an individual’s sleep quality. (1-5, Classification)
\end{itemize}

All tasks is assessed with factors including total sleep time, Steps, mood and other sports data like Burned Calories and Resting Heart Rate over a continuous 14-day period. In terms of range, most tasks are evaluated on a scale of 1-5 or 0-5. A score of 3 represents a normal condition, and 1-2 are scores below normal states, and 4-5 are scores above normal states. For the task of readiness, the scale ranges from 0 to 10, where 0 reflects no readiness for physical activity, and 10 indicates high preparation for exercise. 

The \textbf{label distribution} for each task in this dataset is shown as below:
 
 \begin{figure}[H]
    \centering
    \includegraphics[width=1\linewidth]{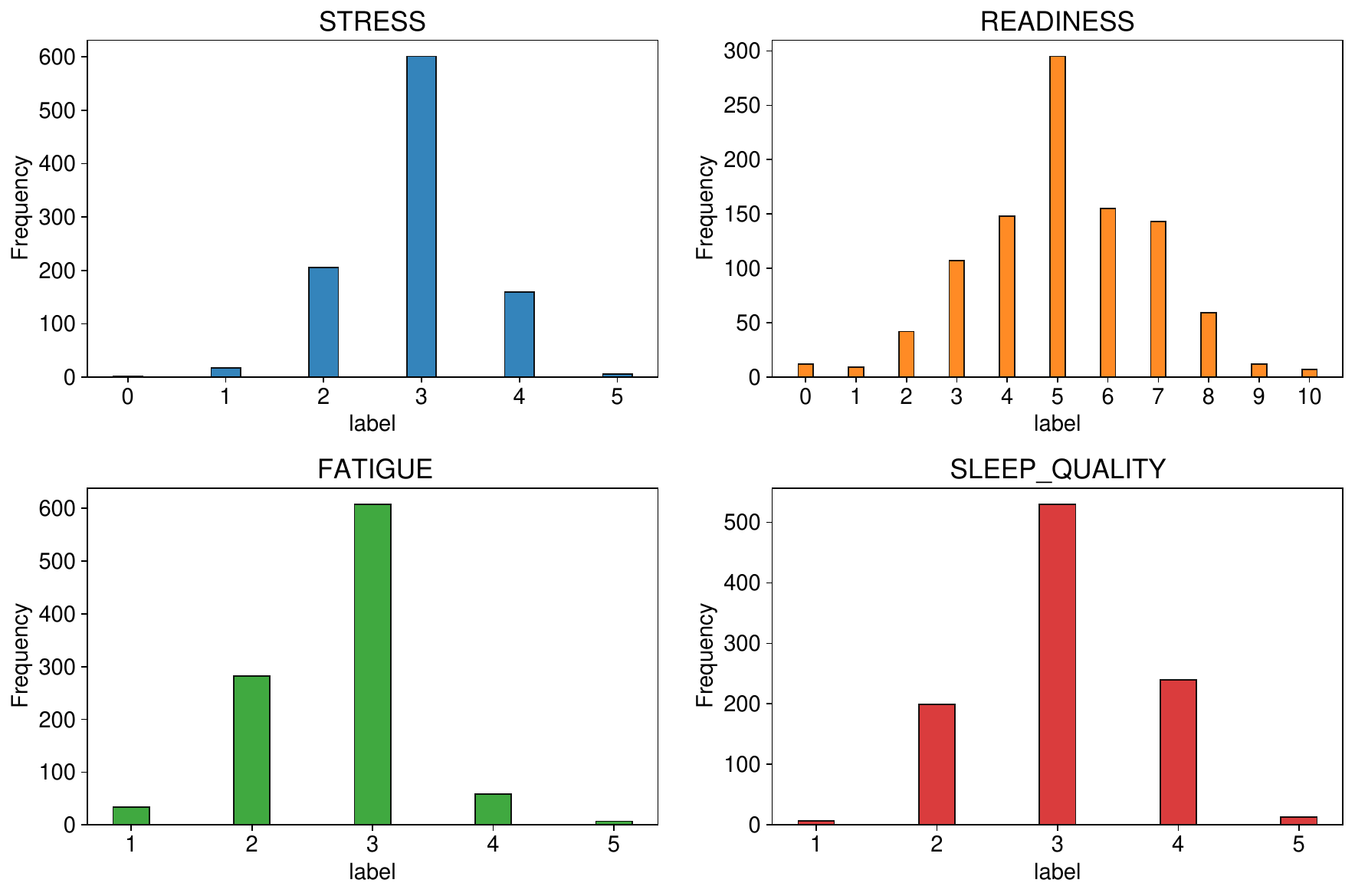}
    \caption{The label distribution of the four tasks in PMData}
\end{figure}

\subsection{GLOBEM}

\begin{itemize}[leftmargin=*]
\item Depression (DEP): estimation of a depression score that analyzes patterns in user's sleeping behavior and activity levels. (0–4, Classification)
\item Anxiety (ANX): estimation of an anxiety score that relies on behavioral markers such as irregular sleep patterns or heightened physiological responses, e.g. increased heart rate, reduced activity levels, and increased sleep disturbances (0–4, Classification)
\end{itemize}

Both the two tasks are assessed on the average of daily steps, sleep efficiency, duration the user stayed in bed after waking up, duration the user spent to sleep, duration the user stayed awake but still in bed, and duration the user spent to fall asleep in the last 14 days. A value of 0 implies the disorder is not present, while a value of 4 indicates severe disorder. Any values between 0 and 4 denote their severity accordingly, such as a value of 1 indicates mild disorder, 2 refers to moderate, and 3 refers to Moderately Severe. 

The \textbf{label distribution} for each task in this dataset is shown below:

\begin{figure}[H]
    \centering
    \includegraphics[width=1\linewidth]{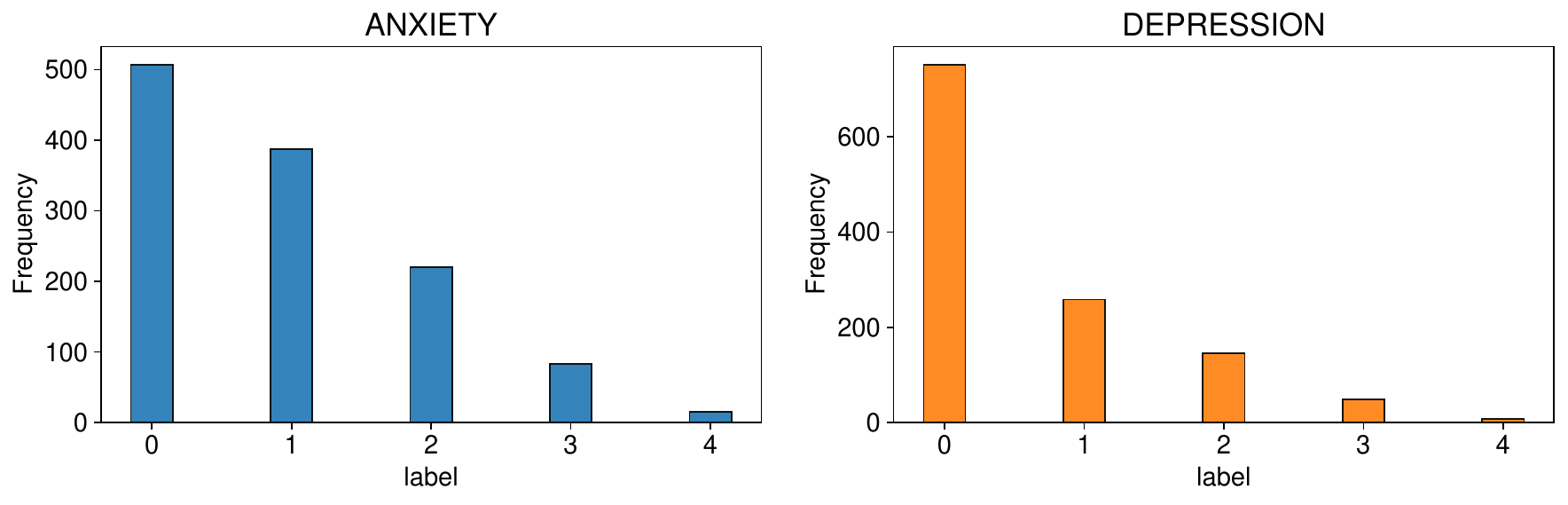}
    \caption{The label distribution of the two tasks in GLOBEM}
\end{figure}

\subsection{AW\_FB}

\begin{itemize}[leftmargin=*]
\item Activity (ACT): estimation of individual's activity intensity type based on sensor data. (0-5, Classification)

\item Calories (CAL): estimation of burned calories that are expended by an individual during physical activities. (no constraint, Regression)
\end{itemize}

Activity is predicted by Steps, Burned Calories, and Heart Rate obtained during an activity period. This label ranges from 0 to 5, corresponding to Self Pace Walk, Sitting, Lying, Running 7 METs, Running 5 METs, and Running 3 METs respectively. Calories are calculated based on Steps, Heart Rate, Duration, Activity Type, and MET Value, where a higher value indicates greater energy expenditure. 

The \textbf{label distribution} for each task in this dataset is shown below:

\begin{figure}[H]
    \centering
    \includegraphics[width=1\linewidth]{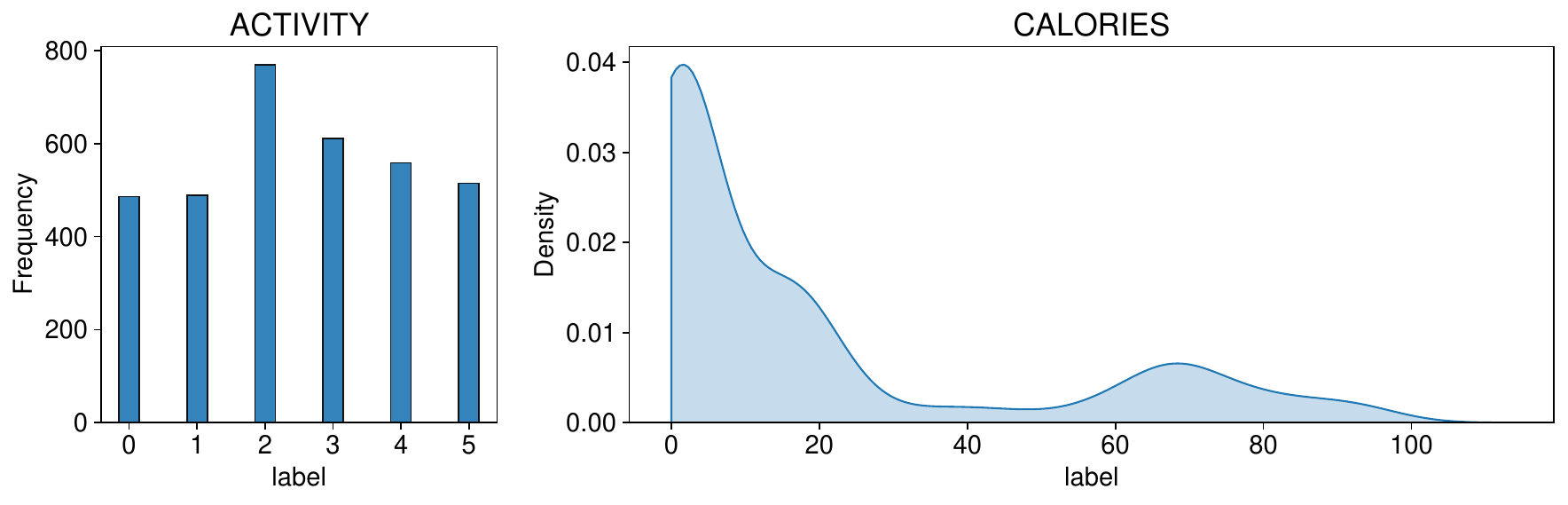}
    \caption{The label distribution of the two tasks in AW\_FB}
\end{figure}

\clearpage
\section{Evaluation Metrics}
\label{Evaluation Metrics}
\subsection{Performance Evaluation}
For SLMs performance evaluation, Mean Absolute Error (MAE) and Accuracy are utilized to assess model prediction performance on health event prediction. 

\textbf{Accuracy (\%)}~\cite{bishop2006pattern} measures the proportion of correctly predicted instances out of all instances. It provides an overview of whether a model performed well overall, with higher values indicating better performance. However, accuracy does not capture the severity or magnitude of errors in misclassified cases, as all errors are treated equally.

\textbf{Mean Absolute Error (MAE)}~\cite{hastie2009elements} quantifies the average magnitude of prediction errors by computing the absolute difference between predicted and actual values. Lower MAE indicates better alignment with ground truth. Unlike Accuracy, which only reflects correctness, MAE distinguishes between small and large errors. For example, predicting ``3'' when the true label is ``4'' yields an error of 1, while predicting ``3'' when the true label is ``10'' yields an error of 7. Thus, MAE captures not only whether predictions are correct but also how close incorrect predictions are to the true values.

In health event prediction, we used both Accuracy and MAE to provide complementary insights. For instance, models that achieve slightly lower accuracy but maintain consistently low MAE may be preferable, as they deliver more reliable outputs than those with higher accuracy but large error magnitudes.

\subsection{Efficiency and Utilization Evaluation}
To further evaluate the efficiency and the actual latency in real cases, all state-of-the-art (SOTA) SLMs that show strong promise will be deployed in processing healthcare field data on a real iPhone 15 Pro Max. To better demonstrate the importance of efficiency on mobile devices, the widely used LLM, Llama 2, is selected and serves as a comparison to the fine-tuned SLMs.
The following metrics suggested by MobileAIBench~\cite{murthy2023} are adopted to evaluate both efficiency and utilization:
\begin{itemize}[leftmargin=*]
\item \textbf{Time-to-First-Token} (TTFT, sec): TTFT is defined by the time of the first token generated to respond to the prompt. It primarily assesses latency, where a lower TTFT indicates a faster response time, allowing users to perceive quicker feedback from the SLM.
\item \textbf{Input Token Per Second} (ITPS, tokens/sec): ITPS is defined by the number of input tokens being processed per second, which refers to how fast the model can read and understand the prompts.
\item \textbf{Output Token Per Second} (OTPS, tokens/sec): OTPS is defined by the number of tokens produced per second after starting to produce tokens, which refers to how fast the model can produce the answer, and access the inference speed. A higher value indicates higher efficiency.
\item \textbf{Output Evaluation Time} (OET, sec): The time model takes to complete a response - assess the overall efficiency of generating an entire response.  A lower value indicates higher efficiency.
\item \textbf{Total Time}: The total time it takes to produce a complete response after receiving a prompt is a comprehensive efficiency metric for how long a model takes to complete a given task from start to finish. A lower value indicates higher efficiency.
\item \textbf{CPU (\%)}: An amount of computational resources used in the inference process.
\item \textbf{RAM (GB)}: An amount of memory needed to run a model during the inference process.
\end{itemize}

During batch evaluations, latency metrics such as TTFT, ITPS, OTPS, OET, and Total Time are calculated as the average time spent or average token processed/generated over a sample size of $N$ (we used 10). CPU utilization is measured by the average load per second during inference, while RAM usage is reported as the maximum memory allocated to the device when the model is running.

\section{SLMs Prediction Distribution}
\subsection{Few-shot Distribution}
\label{Few-shot Distribution}
\begin{figure}[H]
    \centering
    \includegraphics[width=1\linewidth]{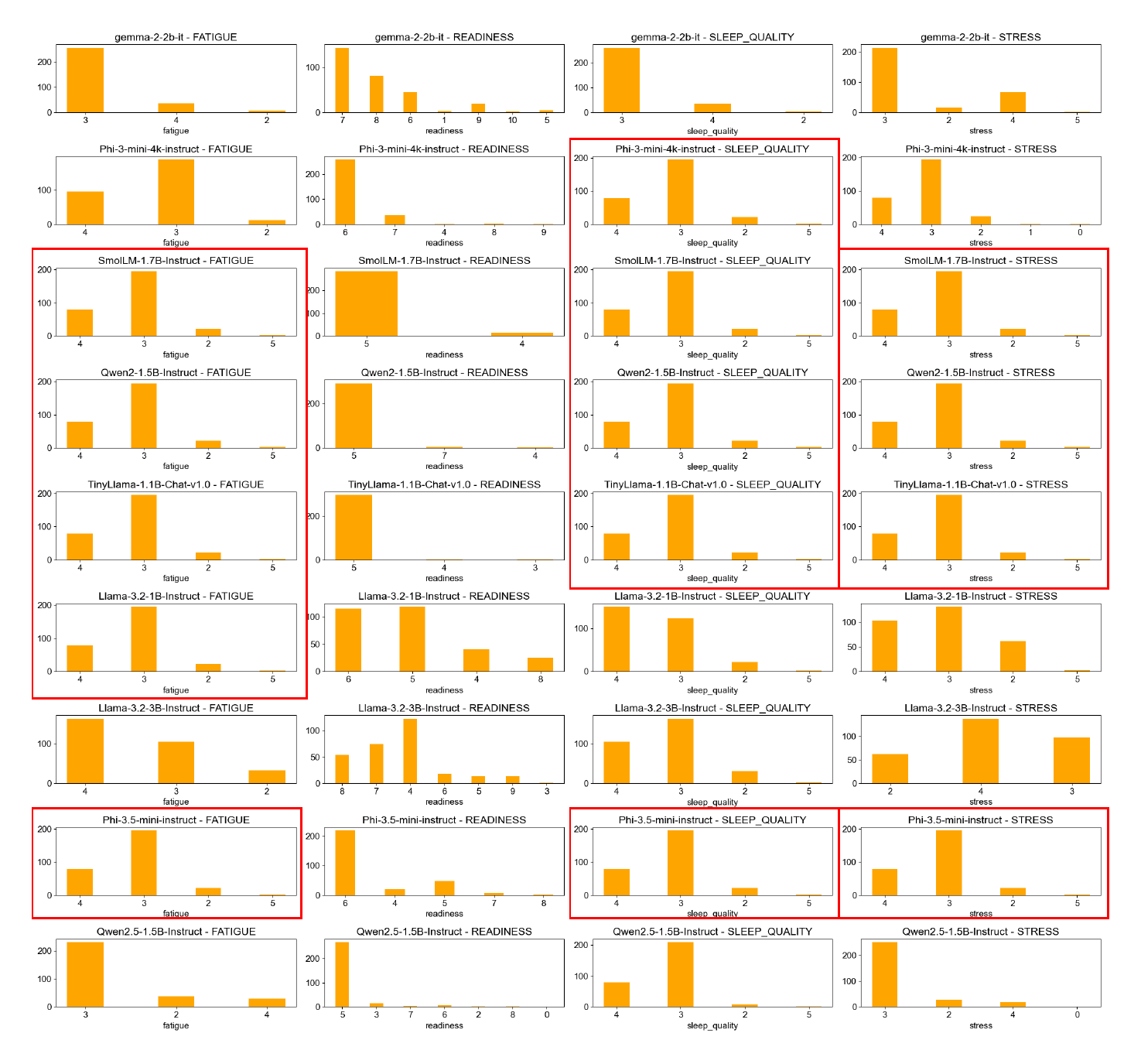}
    \caption{Distribution of predictions for the four tasks in PMData under FS setting}
    \label{fig: Distribution of predictions for the four tasks in PMData under FS setting}
\end{figure}

\clearpage
\subsection{Instructional tuning (LoRA) Distribution}
\label{Instructional tuning (LoRA) Distribution}
\begin{figure}[h]
    \centering
    \includegraphics[width=1\linewidth]{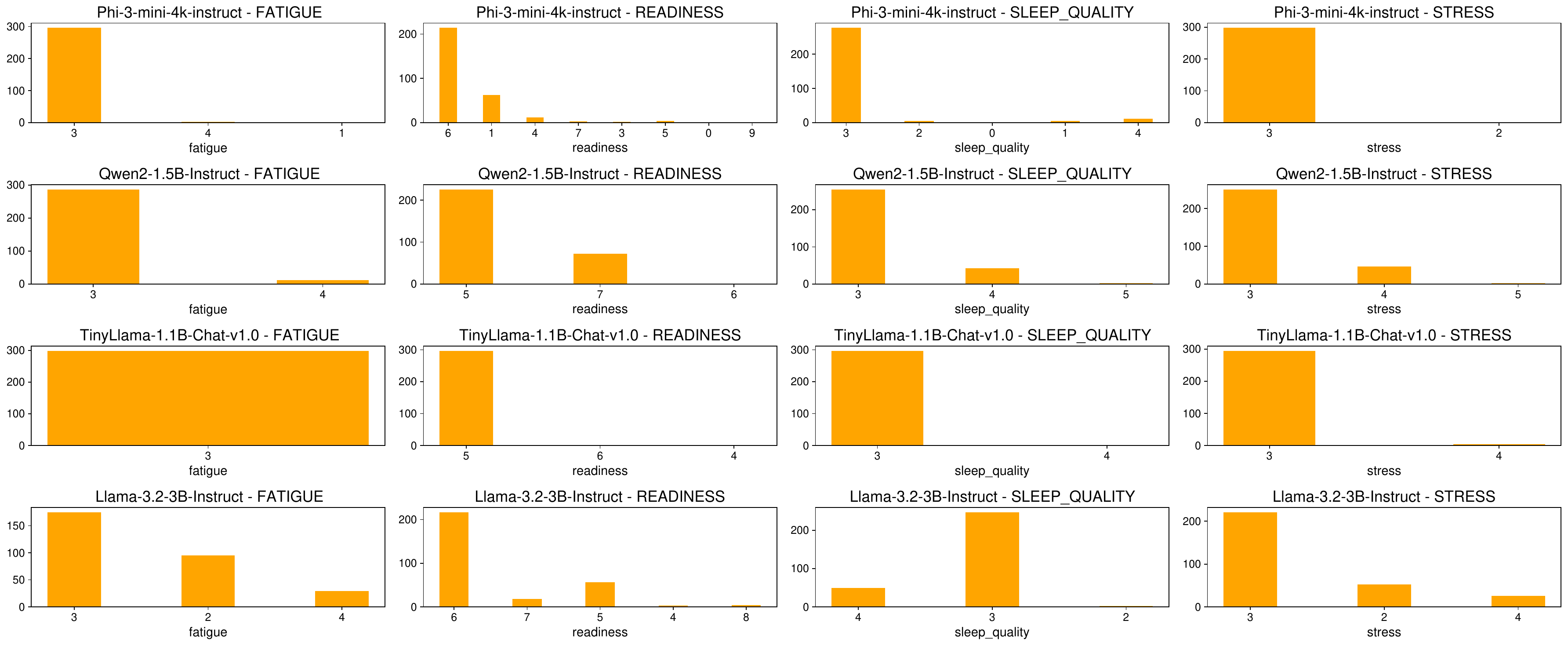}
    \caption{Distribution of predictions for the four tasks in PMData}
\end{figure}

\begin{figure}[h]
    \centering
    \includegraphics[width = 0.8\linewidth]{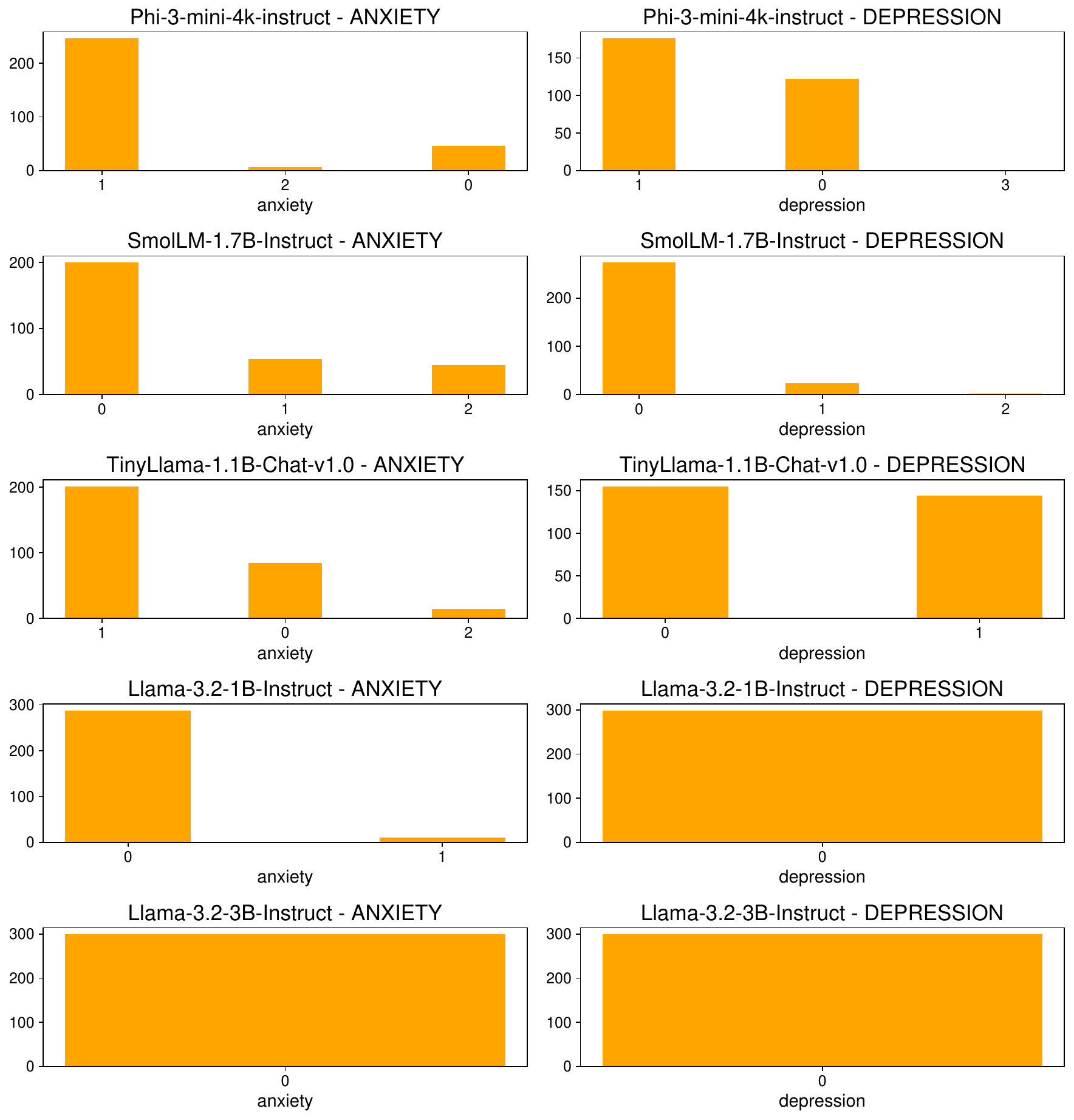}
    \caption{Distribution of predictions for the two tasks in GLOBEM}
\end{figure}

\clearpage
\section{Data Augmentation}

\begin{figure}[h]
    \centering
    \includegraphics[width=1\linewidth]{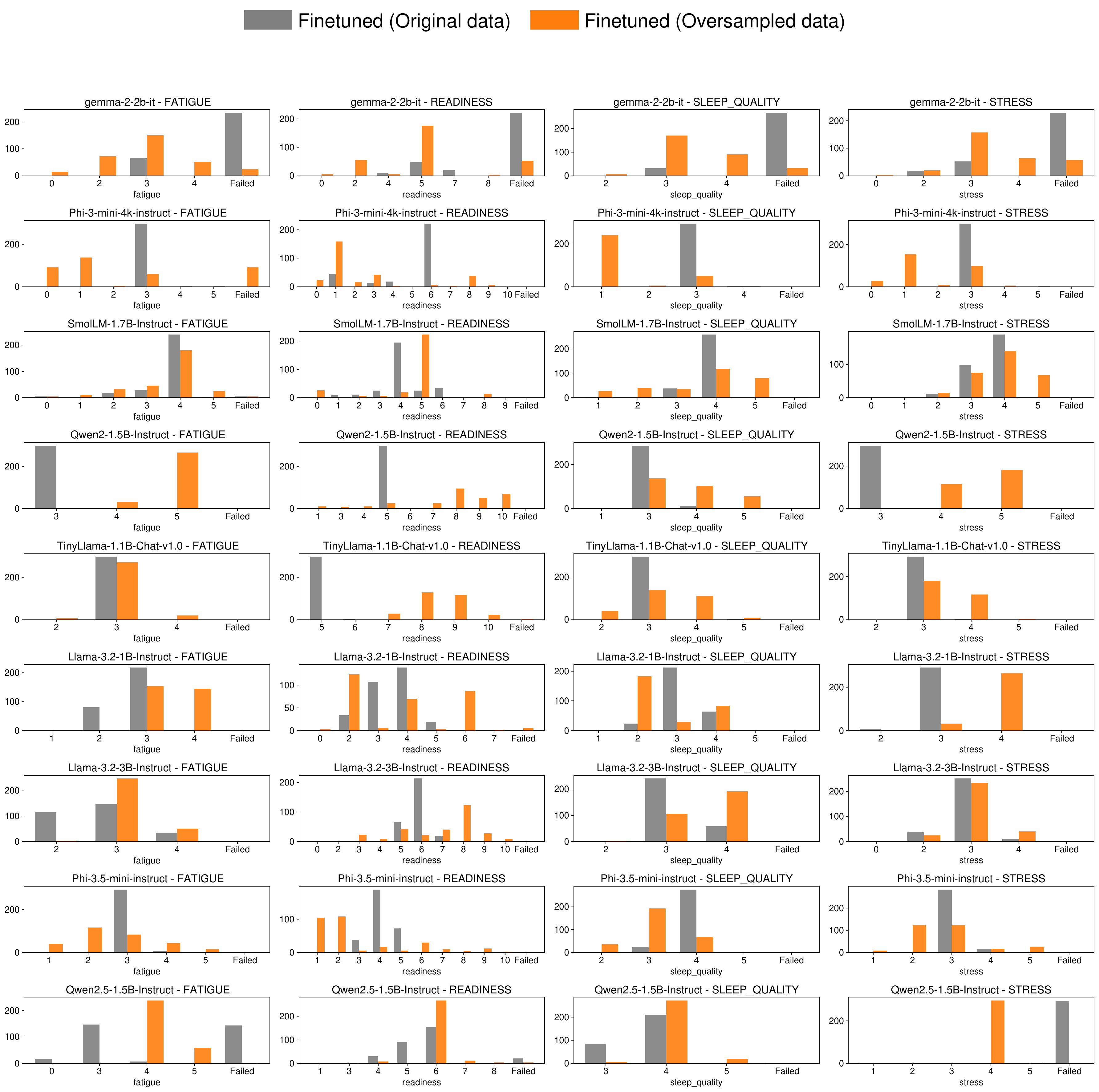}
    \caption{Predicted distributions of SLMs (LoRA) on PMData, comparing models fine-tuned on the original and oversampled datasets.}
    \label{fig: Predicted distributions of SLMs (LoRA) on PMData, comparing models fine-tuned on the original and oversampled datasets.}
\end{figure}


\begin{figure}[h]
    \centering
    \includegraphics[height=0.9\textheight]{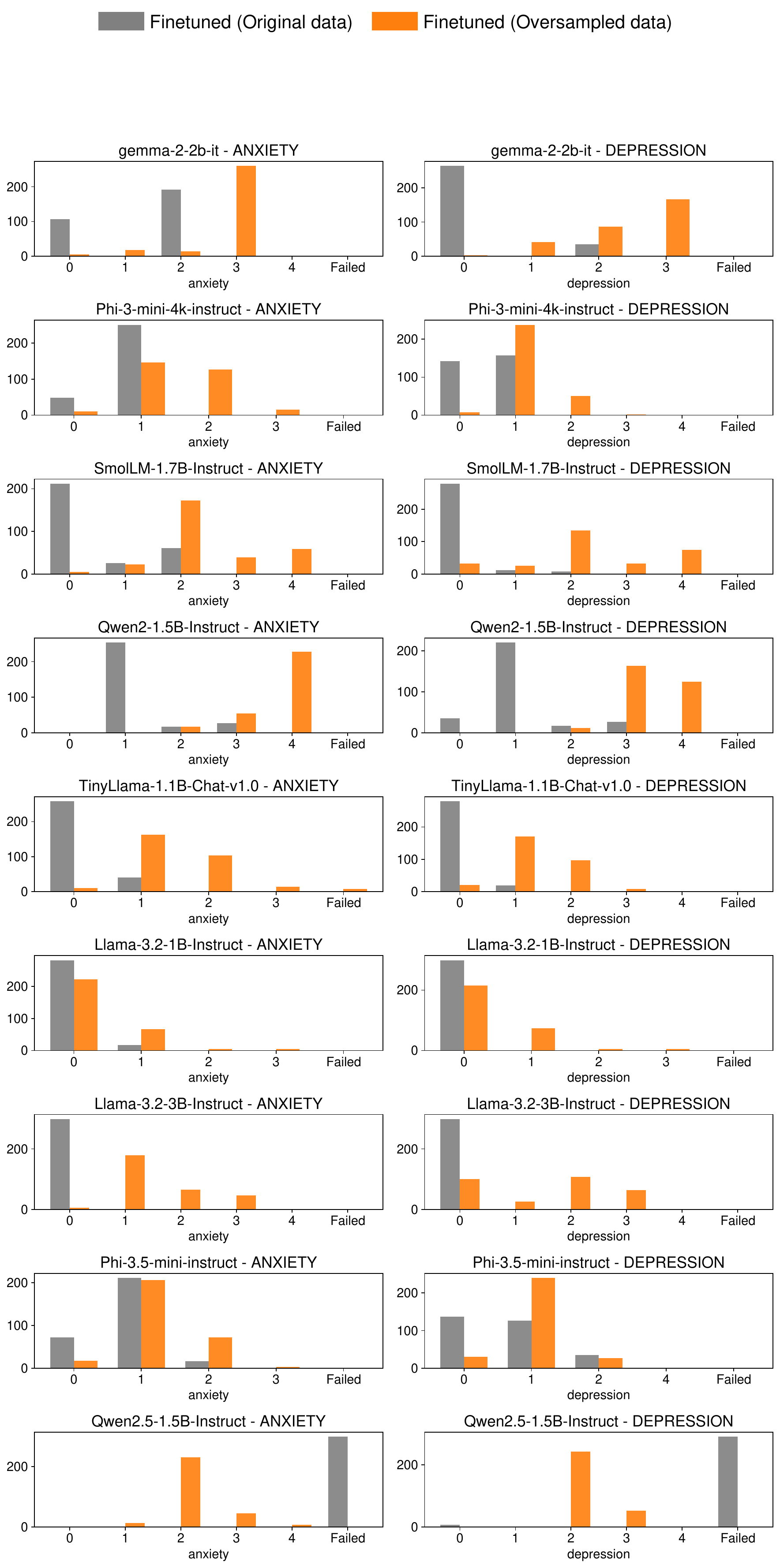}
    \caption{Predicted distributions of SLMs (LoRA) on GLOBEM, comparing models fine-tuned on the original and oversampled datasets.}
    \label{fig: Predicted distributions of SLMs (LoRA) on GLOBEM, comparing models fine-tuned on the original and oversampled datasets.}
\end{figure}

\medskip

\end{document}